\documentclass[10pt,twocolumn,letterpaper]{article}

\usepackage[dvipsnames]{xcolor}

\usepackage{times}
\usepackage{soul}
\usepackage{url}
\usepackage[utf8]{inputenc}

\usepackage{makecell}
\usepackage{amssymb}
\usepackage{enumitem}
\usepackage{colortbl}
\definecolor{lightgray}{gray}{0.95} 
\usepackage{xcolor}
\usepackage{arydshln}

\usepackage[pagenumbers]{cvpr} 

\usepackage[dvipsnames]{xcolor}

\usepackage{times}
\usepackage{soul}
\usepackage{url}
\usepackage[utf8]{inputenc}

\usepackage{makecell}
\usepackage{amssymb}
\usepackage{enumitem}
\usepackage{colortbl}
\definecolor{lightgray}{gray}{0.95} 
\usepackage{xcolor}
\usepackage{arydshln}

\definecolor{cvprblue}{rgb}{0.21,0.49,0.74}
\usepackage[pagebackref,breaklinks,color links,citecolor=cvprblue]{hyperref}

\def\paperID{*****} 
\def\confName{CVPR}
\def\confYear{2024}

\title{The New Agronomists: Language Models are Experts in Crop Management}

\author{Jing Wu\thanks{These authors contributed equally to this work.} \and
        Zhixin Lai\protect\footnotemark[1] \and
        Suiyao Chen\protect\footnotemark[1] \and
        Ran Tao \and
        Pan Zhao \and
        Naira Hovakimyan}

\begin{document}
\maketitle
\begin{abstract}
Crop management plays a crucial role in determining crop yield, economic profitability, and environmental sustainability. Despite the availability of management guidelines, optimizing these practices remains a complex and multifaceted challenge. In response, previous studies have explored using reinforcement learning with crop simulators, typically employing simple neural-network-based reinforcement learning (RL) agents. Building on this foundation, this paper introduces a more advanced intelligent crop management system. This system uniquely combines RL, a language model (LM), and crop simulations facilitated by the Decision Support System for Agrotechnology Transfer (DSSAT). We utilize deep RL, specifically a deep Q-network, to train management policies that process numerous state variables from the simulator as observations. A novel aspect of our approach is the conversion of these state variables into more informative language, facilitating the language model's capacity to understand states and explore optimal management practices. The empirical results reveal that the LM exhibits superior learning capabilities. Through simulation experiments with maize crops in Florida (US) and Zaragoza (Spain), the LM not only achieves state-of-the-art performance under various evaluation metrics but also demonstrates a remarkable improvement of over 49\% in economic profit, coupled with reduced environmental impact when compared to baseline methods. Our code is available at \url{https://github.com/jingwu6/LM_AG}.

\end{abstract}

\section{Introduction}

In today's agricultural landscape, addressing food security and sustainable farming practices is crucial, aligning with the United Nations' goal of Zero Hunger. The challenge of boosting food production for a global population expected to reach 9.6 billion by 2050, while minimizing negative environmental impacts like ecosystem degradation and greenhouse gas emissions, is paramount \cite{searchinger2019creating}. Key factors in crop management, particularly fertilization with nitrogen (N) and irrigation with water (W), significantly affect crop yields and environmental health \cite{reddy2003crop}. However, the previous best practices for these management aspects, derived from empirical experience and academic research \cite{wright2022field,skhiri2012impact}, face uncertainty in their effectiveness against changing climate and market conditions. Therefore, the adequacy of current strategies is questionable, highlighting a need for innovative, efficient, and adaptable management systems. These systems should be capable of devising optimal strategies suitable for varying conditions and objectives, such as maximizing economic profit \cite{ara2021application}. This research is anchored in this context, leveraging advanced AI methods to improve agricultural practices and tackle these critical challenges.

Reinforcement learning (RL) has shown exceptional capabilities in tasks that involve sequential decision-making (SDM), such as in robotics, gaming and multi-agent systems \cite{kaufmann2018droneracing-rl,mnih2015human-deepRL, liu2022distributed, liu2023meta, liu2024learning,cheng2023safe,cheng2022improving}. This success suggests a significant potential for RL in optimizing crop management, which at its core is an SDM problem. Given the need for numerous interactions between the RL agent and the environment during policy training, field trial-based methods are impractical. Consequently, the use of crop models to simulate both the crop and its environment, providing a platform for interaction with the RL agent, appears to be the most feasible approach \cite{palmer2013influence,attonaty1997using}.

\begin{figure*}[t]
  \centering
  \includegraphics[width=0.7\linewidth]{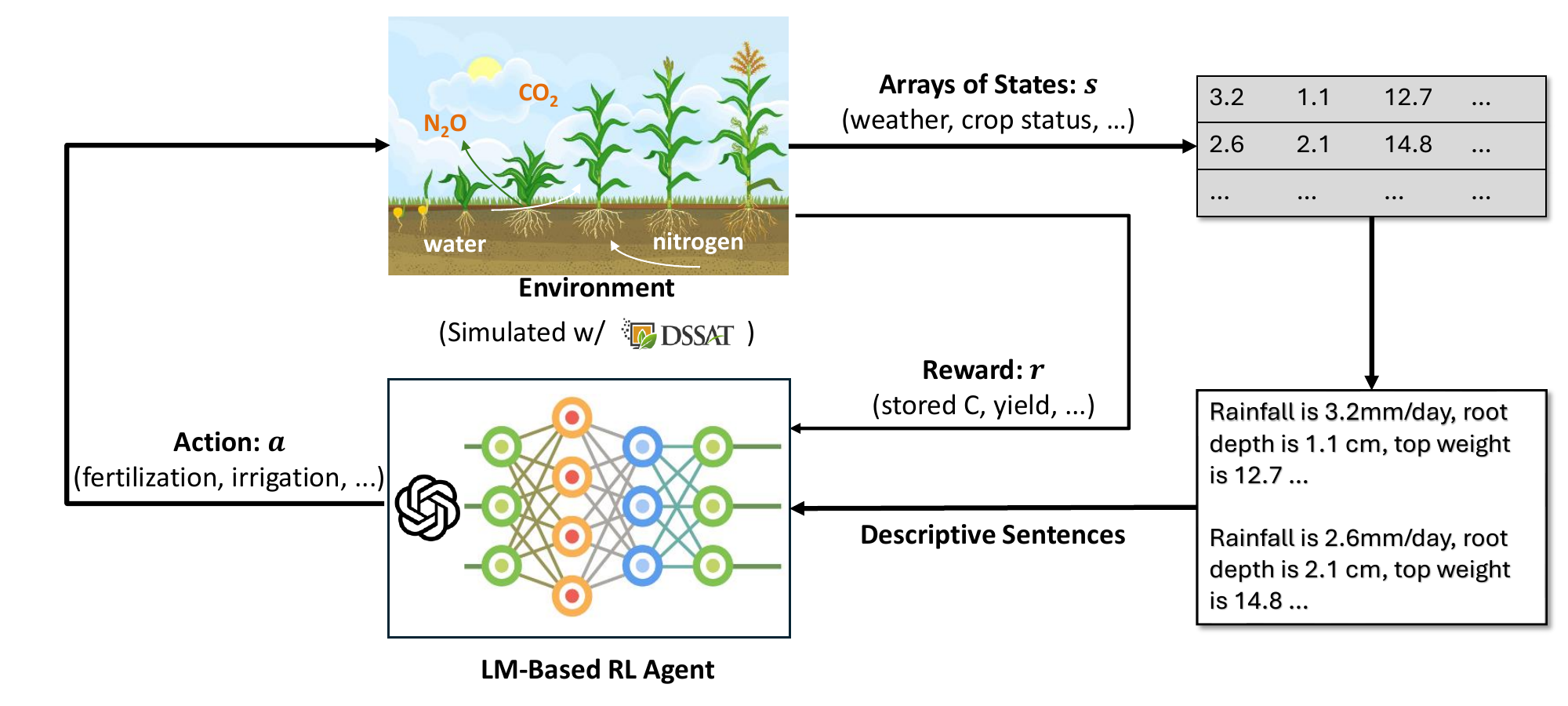}
   \caption{Framework and pipeline of the intelligent crop management system using LM-based RL}
   \label{fig:Framework}
\vspace{-5mm}
\end{figure*}

Recently, the authors of \cite{wu2022optimizing,tao2022optimizing} proposed to train management policies for crop management using deep RL with DSSAT \cite{jones2003dssat} and Gym-DSSAT \cite{romain2022gym}, one of the most widely used crop models in the world. Their trained policies, both under full and partial observations, outperformed baseline policies by achieving higher yields or similar yields with reduced nitrogen (N) fertilizer input. However, there are limitations to these approaches. Firstly, the models primarily employed Multilayer Perceptrons (MLPs), which, while effective, have limited fitting power compared to more complex architectures. This limitation could potentially constrain the models' ability to capture the intricate dynamics of crop growth and management fully. Secondly, the reliance on MLPs limits the incorporation of additional descriptive features for state representations in the model. These features could include various environmental, soil, and crop growth parameters that are crucial for precise agricultural decision-making. On one hand, the path to further improve the existing framework could be replacing MLPs with more sophisticated tabular learning models \cite{wu2024switchtab,chen2023recontab}, enhancing the training process \cite{wang2023balanced, zhang2024trade, hu2023many, wu2023hallucination} and integrating domain knowledge and theoretical assumptions for model design \cite{ding2019confidence, chen2017multi, wu2023extended} and uncertainty quantification \cite{chen2023quantifying, chen2020optimal, bingjie2023optimal, chen2017personalized}, which has been effective in many traditional applications \cite{liu2024particle, chen2019claims, chen2018data, wu2023genco, shi2017combining}. One the other hand, Language Models (LMs) have demonstrated great potentials in handling complex information and providing insights for better decisions, with generation \cite{lai2024adaptive,li-etal-2023-synthetic} and reasoning \cite{liu2024temperature, su2024large} powers and broad applications \cite{li2023joyful,liu2024news,zhou2023thread}. This gap in the model's design raises a critical question: Can LMs serve as viable alternatives for RL agents in these crop management tasks? The limitations of the existing models and this pivotal question motivate the present paper.

In this paper, we present an intelligent crop management framework, depicted in Figure \ref{fig:Framework}, that incorporates a powerful LM, and crop simulations via DSSAT and Gym-DSSAT. Concretely, we transform the states from simulation tools, typically arrays of numbers, into more descriptive sentences. This conversion enables a significant shift in our approach: we replace the traditional MLP-based RL agent with an LM-based RL agent. This new agent leverages LMs to encode these descriptive state sentences into embeddings, thereby capturing a more informative and nuanced understanding of the states. Meanwhile, we notice that LMs have shown distinctive cognitive capabilities, which include advanced thinking \cite{wei2022chain, chen2023you}, robust memory functions \cite{park2023generative}, reflective skills \cite{shinn2024reflexion, li2022tip}, as well as multi-modal capabilities \cite{zhou2024visual, xin2023mmap, wang2023emp}. As a result, the RL agent should be equipped with the ability to comprehend complex aspects of crop growth and simulation environments. We, therefore, anticipate that the incorporation of LMs will markedly improve the performance of the RL agent in crop management tasks.

To demonstrate the effectiveness of the proposed method, we conducted the case studies simulating maize crops management in Florida, USA, and Zaragoza, Spain. This choice of locations aligns with the settings used in previous studies \cite{tao2022optimizing}. In both scenarios, the policies trained by our framework exhibited superior performance compared to the previous state-of-the-art approaches; the baseline was derived from either maize production guidelines recommended by agricultural experts as well as survey results on actual management practices of maize farmers. Additionally, continuing in the vein of established research, our investigation also includes the training of RL-based policies with well-recognized reward functions \cite{tao2022optimizing}. These functions are designed to represent different balances among key factors: crop yield, resource utilization, and environmental impact, particularly focusing on nitrate leaching during the crop growth cycle. 
In summary, the primary contributions of our work can be delineated as follows:

\begin{itemize}[noitemsep,topsep=0pt]
    \item [$\bullet$] We investigate a critical yet under-explored question: Can LMs serve as better alternatives for RL agents in crop management tasks to offer more nuanced and effective solutions and advance the state of intelligent crop management systems?
    
    \item [$\bullet$] To the best of our knowledge,  this work marks the first attempt to integrate descriptive language to represent agricultural states and to employ LMs in the pursuit of optimal crop management policies.

    \item [$\bullet$] We empirically demonstrate that our proposed framework exceeds the performance of existing state-of-the-art approaches in various key aspects, including crop yield, resource utilization, and environmental impact.

\end{itemize}

\section{Related Work}
\subsection{Crop Management with RL}
Considering crop management as a Markov decision process (MDP), initial efforts in applying reinforcement learning (RL) to derive optimal management strategies from simulators have emerged, but this field is still developing. An early attempt to use a basic RL approach for wheat management in France was documented by \cite{garcia1999rl-wheat}. Another study by \cite{sun2017rl-irrigation} focused on optimizing irrigation for maize in Texas, US. However, these studies had limitations, such as narrow state and action spaces. For example, \cite{sun2017rl-irrigation} included only a single state variable for RL training. In \cite{ashcraft2021ml-aided}, the researchers applied the proximal policy optimization (PPO) algorithm for fertilizer and irrigation policy optimization, but their results did not significantly surpass existing baseline methods in simulations. More comprehensive research in this area is represented by \cite{wu2022optimizing}, which examined N fertilization for maize in Florida and Iowa. Subsequent studies have expanded on this approach using different crop models \cite{kallenberg2023nitrogen,madondo2023swat}. Addressing the challenges of partially observed crop management, authors of \cite{tao2022optimizing} explored the use of imitation learning, training an RL agent with a broad set of state features and then applying it to a subset of these features.

\subsection{Crop Models for RL}
\label{section:dssat}

The necessity for crop models arises from the practical challenges of conducting real-world farming experiments, which are often laborious, time-consuming, and expensive. These models are crucial for assessing the impact of climate change and various management practices on crop production \cite{zhao2019simple}. Among the numerous crop simulation models developed, APSIM and DSSAT are particularly notable for their widespread use and continuous updates, providing accurate estimations of crop production in relation to multiple factors \cite{jones2017brief,bassu2014various,asseng2013uncertainty,tao2022optimizing,wu2022optimizing}. Traditional crop models, however, typically require the pre-definition of management practices before simulations, a limitation when compared to the dynamic decision-making capability of reinforcement learning (RL). To bridge this gap, efforts have been made to integrate RL with crop models, enabling real-time decision-making during simulations. Innovations like the CropGym environment and interfaces based on the SIMPLE crop model for russet potatoes, both utilizing the Open AI Gym framework \cite{brockman2016openai,overweg2021cropgym,zhao2019simple}, demonstrate the feasibility of RL in crop management. Despite this progress, some of these models oversimplify essential crop and environmental details. In contrast, Gym-DSSAT, built on the robust DSSAT model, allows for detailed, daily interactions between the RL agent and the simulated environment, a significant advancement in optimizing nitrogen and irrigation management \cite{romain2022gym,wu2022optimizing,tao2022optimizing}.

\subsection{LM for Decision-making}

In recent years, there has been a surge in studies utilizing pre-trained LMs as decision-making agents. These models' remarkable capabilities have been harnessed across various domains, generating improved control plans for diverse robots and agents \cite{huang2022language,huang2022inner,raman2022planning,mees2023grounding,chen2023open,ahn2022can,liang2023code}. Notably, researchers of \cite{kim2024language} developed LM-based agents for user interface (UI) interactions, while ReAct \cite{yao2022react} integrated action decisions with natural language reasoning, demonstrating promising results.

To the best of our knowledge, our work represents the first endeavor to leverage the LMs in formulating optimal management strategies for crop models in agriculture.

\section{Methods}

\subsection{Problem Formulation}
In this study, we approach nitrogen fertilization and irrigation management as a finite Markov Decision Process (MDP), following the paradigm of previous work \cite{tao2022optimizing,wu2022optimizing}. Each day, denoted as day $t$, involves the agent receiving the environmental state, $s_t$, and subsequently selecting an action $a_t$ from the action space $\mathcal{A}$. This selection is guided by a policy $\pi(s_t,\theta_t)$, where $\theta_t$ symbolizes the policy parameters on that particular day, and notably, the policy in this context is a pretrained language model. The state $s_t$ encompasses vital data pertaining to weather, plant growth, and soil conditions, as simulated for that day. The action $a_t$ is composed of two key decisions: the quantity of nitrogen fertilizer, denoted as $N_t$, and the amount of irrigation water, $W_t$, to be applied. The effectiveness of these decisions is quantified by the reward $r_t(s_t,a_t)$, which is calculated based on the outcomes of $s_t$ and $a_t$, defined as:

\begin{equation}\label{eq:reward}
    r_t(s_t,a_t) \!=\! \left\{\hspace{-2mm}
    \begin{array}{ll}
         w_1Y\!-\! w_2 N_t \!-\! w_3 W_t \! - \! w_4 N_{l,t} \! & \hspace{-2mm}\textup{if }  \textup{harvest at $t$,}    \\
        \!- w_2 N_t \!-\! w_3 W_t \! - \! w_4 N_{l,t}  & \hspace{-2mm} \textup{otherwise,} 
    \end{array}
    \right.
\end{equation}
where $w_1, w_2, w_3, w_4, Y, N_{l,t}$ denote four custom weight factors, yield at harvest and the amount of nitrate leaching on a given day, respectively. Both $Y$ and $N_{l,t}$ are derived from the state variable $s_t$. The design of the reward function, characterized by the weights $w_1, w_2, w_3, w_4$, is pivotal in steering the agent's strategy. The agent's objective is to identify the optimal policy $\pi(s_t,\theta_t)$ that selects action $a_t$ to maximize the total future discounted return. This return, defined as $R_t = \sum_{\tau=t}^T \gamma^{\tau-t} r_\tau$, captures the accumulated reward from the current action $a_t$ to the future rewards, discounted by factor $\gamma$.

\subsection{LM-based RL Agent}
To harness the full potential of language models (LM) in comprehending crop models and identifying optimal management strategies, we made adaptations to the state variables from the simulation tool, specifically Gym-DSSAT. Traditionally, the state in such simulations is represented by an array of variables reflecting various crop and environmental conditions, like rainfall and root depth. However, this format does not provide a direct correlation between the variables and their descriptive meanings, posing a challenge for RL agents to interpret each variable independently. To overcome this, we transformed the raw data into a more language-friendly format. Each variable name and its corresponding value were combined into coherent sentences. This approach essentially transforms the state data into a format that is more accessible and interpretable by LMs, allowing for a more intuitive and efficient exploration of management practices.

In our approach, we have innovated by substituting the traditional MLPs with a distilled and pre-trained BERT model from \cite{sanh2019distilbert} serving as the RL agent. This advanced model is utilized to encode the concatenated sentences, which represent the state variables, into feature embeddings. Following this encoding process, we introduce a few fully connected layers connected to the distilled BERT encoder. These layers are responsible for transforming the generated feature embeddings into a format that aligns with the action space of the RL agent. This novel architecture not only leverages the linguistic understanding of BERT but also ensures that the complex relationships within the crop management data are effectively captured and translated into actionable insights.

\subsection{Policy Training with LM}
In this study, we use the Deep Q-Network (DQN) from \cite{mnih2015human-deepRL} to train our agent. The goal is to learn an optimal policy that maximizes the future discounted return, denoted as $R_t$. A novel aspect of our approach is the integration of the distilled BERT model to represent the action-value function, also known as the Q function, within the DQN framework. This Q function, formally defined as $Q^\pi(s,a) = \mathop{\mathbb{E}}[R_t|s_t=s,a_t=a,\pi]$, is essential for calculating the expected future discounted return from state $s$ when action $a$ is taken, following policy $\pi$.

The objective is to refine the parameters of the Q-network to pinpoint the optimal Q function, $Q^\star(s,a)$, which indicates the highest return possible from state $s$ by taking action $a$ and adhering to the optimal policy. For selecting the optimal action in state $s_t$, we employ a greedy policy defined as $a_t^\star = \max_{a\in \mathcal{A}}Q^\star(s_t,a)$. Training the Q-network, which effectively means training the policy, involves minimizing the following loss function:
\begin{equation}
L_i(\theta_i) \! \triangleq
  \!\! \mathop{\!\mathbb{E}}_{(s,a,r,s')}\!\left[r\!+\!\gamma \max_{a'\in \mathcal{A}}Q(s'\!,a';\theta_i^{-}) \!-\! Q(s,a;\theta_i)\!\right]\!.
\end{equation}
Here, $s,a,r,s'$ denote the state, action, reward, and next state, respectively, with $\gamma$ representing the discount factor, and $\theta_i^{-}$ representing the parameters of a target network defined earlier. The tuples $(s,a,r,s')$ for the loss function are randomly sampled from the replay buffer, a collection of prior state-action-reward-next state tuples accumulated during training.

\subsection{Crop Simulations with Gym-DSSAT}
Similar to \cite{wu2022optimizing,tao2022optimizing}, we leverage Gym-DSSAT \cite{romain2022gym}, a Gym interface for DSSAT that enables the agent to interact with the simulated environment (i.e., reading the weather, soil, and crop information and applying management practices) on a daily basis. For more details about DSSAT and Gym-DSSAT, readers can refer to Section \ref{section:dssat}.

\section{Experiments and Results}
In this section, various experiments are conducted on real-world datasets to demonstrate the effectiveness and superiority of the proposed framework. The experiment settings are introduced in Section \ref{exp: settings}, where the setup and techniques used in the experiments are detailed. Following this, the training and evaluation details are illustrated in Section \ref{exp: details}, providing the necessary details to reproduce the work of the paper. Then, we present the evaluation results, where the performance of our proposed method is compared against existing baselines and SoTA approaches in Section \ref{exp: results}. Lastly, ablation studies are conducted for policy training in Section \ref{exp: ablation}.

\subsection{Experimental Setup}
\label{exp: settings}
The experiments focusing on training policies for nitrogen and irrigation management in maize crops were conducted through two distinct case studies, both utilizing real-world data. The first of these case studies was set in a simulated environment replicating Florida, USA, in 1982, while the second case study was based on the simulated conditions of Zaragoza, Spain, in 1995. The primary objective of these case studies was to test and demonstrate the viability and advantages of the proposed framework, rather than preparing it for immediate real-world application. For those interested in the specifics of deploying this framework in practical settings, further details are provided in Section \ref{sec:deployment}.

For each case study, DQN was used to train the LM-based RL agent under full observation. The performance of all trained policies was evaluated in simulation, and compared with baseline policies and previous state-of-the-art methods as mentioned in \cite{tao2022optimizing}. The baseline for the Florida study was based on a maize production guide for Florida farmers \cite{wright2022field}, and for the Zaragoza study it was derived from survey data on maize farming practices in Zaragoza \cite{malik2019dssat,skhiri2012impact}.

The framework was implemented to train the RL agent under full observation. This approach involved testing with five different reward functions, each designed to demonstrate the adaptability of the framework to various trade-offs. These trade-offs include balancing crop yield, N fertilizer use, irrigation water use, and environmental impact. This variety in reward functions allows the framework to be evaluated across a range of scenarios and objectives, showcasing its flexibility in addressing different agricultural management priorities.

\subsection{Implementation Details and Evaluation Metrics}
\label{exp: details}
\textbf{Implementation Details.} 
The RL agent in our study employs a combination of DistilBERT and a three-layer fully connected neural network for feature adaptation. The process begins with DistilBERT encoding the state inputs into 768-dimensional embeddings. Notably, the parameters of DistilBERT are trained end-to-end in this model. After this initial encoding, the embeddings are passed through fully connected layers, one with 512 units and the other with 256 units. The final layer in this sequence is responsible for mapping these processed embeddings to the action space, completing the flow from the input state to the actionable output in the RL framework. The discrete action space is defined as follows:

\begin{align}
\label{eq:action space}
    \mathcal A =\{40k  \frac{\textrm{kg}}{\textrm{ha}} \text{ N fertilizer }\; \&\; 6k \frac{\textrm{L}}{m^2} \textrm{ Irrigation water} \},
\end{align}
where $k=0,1,2,3,4$, resulting in a total of 25 possible actions. This action space design incorporates standard quantities of N fertilizer and irrigation water that are typically applied by farmers in a single day. It also allows for a wide range of options, aiding the discovery of effective policies. The discount factor is meticulously set at 0.99. To facilitate the neural network's updates, Pytorch is employed alongside the Adam optimizer \cite{kingma2014adam}, characterized by an initial learning rate of 1e-5 and a batch size of 512. This setup is strategically chosen to optimize the learning process while ensuring efficient computation.

The direct application of DistilBERT's tokenizer to numerical values introduces significant training instability. Concretely, numerical values are often segmented into multiple tokens, resulting in considerable variance for small numerical differences. For instance, the number 360 tokenizes into [9475], while 361 splits into [4029, 2487], indicating a disproportionate representation of adjacent numbers. This inconsistency can amplify instability during training. Additionally, the tokenization of decimal numbers compounds this issue. For example, 0.1 translates into [1014, 1012, 1015], where `0' and the decimal point are tokenized separately, leading to unnecessary token proliferation and computational inefficiency.

\begin{table}[t]
\small
\centering
\begin{tabular}{l|r|r|r|r|l }\toprule
& \makecell{$w_1$\\ ($Y$)}  & \makecell{$w_2$\\ ($N_t$)} & \makecell{$w_3$ \\($W_t$)} & \makecell{$w_4$ \\($N_{l,t}$)}&Note \\ \midrule
RF 1 & 0.158 & 0.79    & 1.1     & 0 & Economic profit    \\ 
RF 2 & 0.158 & 0.79    & 0       & 0 &  Free water \\ 
RF 3 & 0.158 & 0    & 1.1    & 0 &  Free N fertilizer \\ 
RF 4 & 0.158 & 1.58    & 1.1     & 0  & Doubled N price  \\ 
RF 5 & 0.2   & 1       & 1       & 5  &  With N Leaching  \\ \bottomrule
\end{tabular}
\caption{Weights used in each reward function (RF) defined by  equation \eqref{eq:reward}}
\label{table:weight}
\vspace{-5mm}
\end{table}

To address the tokenization challenges with numerical values in our model, we have developed a straightforward yet effective data preprocessing technique. This method involves normalizing numerical values to fit within the range of [0, 300] and subsequently utilizing only the integer portion for tokenization. The decision to cap the range ensures that each normalized number corresponds to a single token, thereby simplifying and stabilizing the tokenization process. Additionally, focusing solely on the integer part helps to minimize the number of tokens used. We achieve a succinct representation comprising 27 distinct tokens, which includes 25 feature-specific tokens plus two special tokens ([CLS] and [SEP]). This streamlined token set not only improves the stability of the training process but also enhances its computational efficiency, which is crucial for the complex task of crop management optimization using RL and language models.



\textbf{Evaluation Metrics.} In each case study, we employed reward functions in line with the approach described in previous research \cite{tao2022optimizing}. Specifically, five distinct reward functions for $r_t$ derived from Equation \eqref{eq:reward} were utilized to train the RL agent. For each reward function, a single trained policy was selected for evaluation. The parameters for each reward function (RF) are detailed in Table \ref{table:weight}.

RF1 quantifies the economic profit (\$/ha) that farmers accrue, calculated based on the estimated prices of maize and the costs of N fertilizer and irrigation water, as referenced from \cite{mandrini2022exploring} and \cite{wright2022field}. RF2-RF4 represent variations of economic profit under different scenarios. Specifically, RF2 addresses the hypothetical situation where irrigation water is free of cost; RF3 considers the case where N fertilizer is free; and RF4 models a scenario in which the price of N fertilizer is doubled.

In contrast to RF1-RF4, which focus solely on economic profit, RF5 incorporates an additional environmental aspect, specifically nitrate leaching. Nitrate leaching is a significant environmental concern as it contributes to problems like eutrophication of water bodies and soil degradation \cite{di2002nitrate}. RF5 is structured to balance yield, N fertilizer, and irrigation use while assigning a substantially higher weight to nitrate leaching. This approach aims to minimize nitrate leaching while still achieving favorable economic outcomes.

\begin{table*}[ht]
\small
\centering
\begin{tabular}{l rrrrrrrr}
\toprule
Florida Case & \makecell{N Input \\ (kg/ha) $\downarrow$} & \makecell{Irrigation\\ (L/m$^2$) $\downarrow$} & \makecell{Yield\\ (kg/ha) $\uparrow$} &RF1 $\uparrow$ &RF2 $\uparrow$ &RF3 $\uparrow$ &RF4 $\uparrow$ &RF5 $\uparrow$ \\ \midrule
Empirical Baseline         & 360     & 394            & 10772  &984                         &1417                        &1269                        & 700                         & 338
\\  \midrule
Policy1: Traditional Agent      & {200}     & \textbf{120}              & 10852  &{1425}                        &1557                       &1538                       &1267                        & {1673}
\\ 
\rowcolor{lightgray}  Policy1: LM-based Agent (Ours)       & \textbf{122}     & {192}            & \textbf{11402}  &\textbf{1464}                        &\textbf{1675}                       &\textbf{1590}                       &\textbf{1337}                        & \textbf{1748} 
\\  \midrule

Policy2: Traditional Agent        & 200     & 732             & 11244    &813                         & {1619}               &971                        &655                         &1020                    \\ 
\rowcolor{lightgray}  Policy2: LM-based Agent (Ours)       & \textbf{160}     & \textbf{510}            & \textbf{11474}  &\textbf{1126}                        &\textbf{1687}                       &\textbf{1252}                       &\textbf{999}                        & \textbf{1330} 
\\  \midrule

Policy3: Traditional Agent       & 19920     & \textbf{108}               & 10865    &-1.4e4                       &-1.4e4                     & {1598}                      &{-3.0e4}                       &-4.9e4                     \\
\rowcolor{lightgray}  Policy3: LM-based Agent (Ours)       & \textbf{10000}     & {264}            & \textbf{13152}  &\textbf{-6.1e3}                        &\textbf{-5.8e3}                       &\textbf{1788}                       &\textbf{-1.4e4}                        & \textbf{-3.8e4} 
\\  \midrule

Policy4: Traditional Agent       & 160     & 102              & \textbf{10358}  &1398                        &\textbf{1510}                        &1524                        &{1272}                        &1635     \\
\rowcolor{lightgray}  Policy4: LM-based Agent (Ours)       & \textbf{160}     & \textbf{36}            & {10192}  &\textbf{1428}                        &{1468}                       &\textbf{1555}                       &\textbf{1302}                        & \textbf{1647} 
\\  \midrule

Policy5: Traditional Agent       & 200     & 138              & 10926 &1417              &1568                        &1575              &1259               &1651     
\\  

\rowcolor{lightgray}  Policy5: LM-based Agent (Ours)       & \textbf{160}     & \textbf{60}            & \textbf{11280}  &\textbf{1590}                        &\textbf{1656}                       &\textbf{1716}                       &\textbf{1463}                        & \textbf{1841} 

\\ \bottomrule

\end{tabular}
\caption{The evaluation results of our trained policies, comparing them with previous SoTA methods and baseline policies. `Policy x' refers to the policy optimized using the reward function `RF x'. The `RF x' column details the cumulative rewards for each policy, calculated in accordance with `RF x'. Details of each reward function can be found in Table \ref{table:weight}. The best value is highlighted in \textbf{bold}.}
\label{table:TP_results}
\vspace{-5mm}
\end{table*}

\begin{figure}[t]
  \centering
   \includegraphics[width=0.8\linewidth]{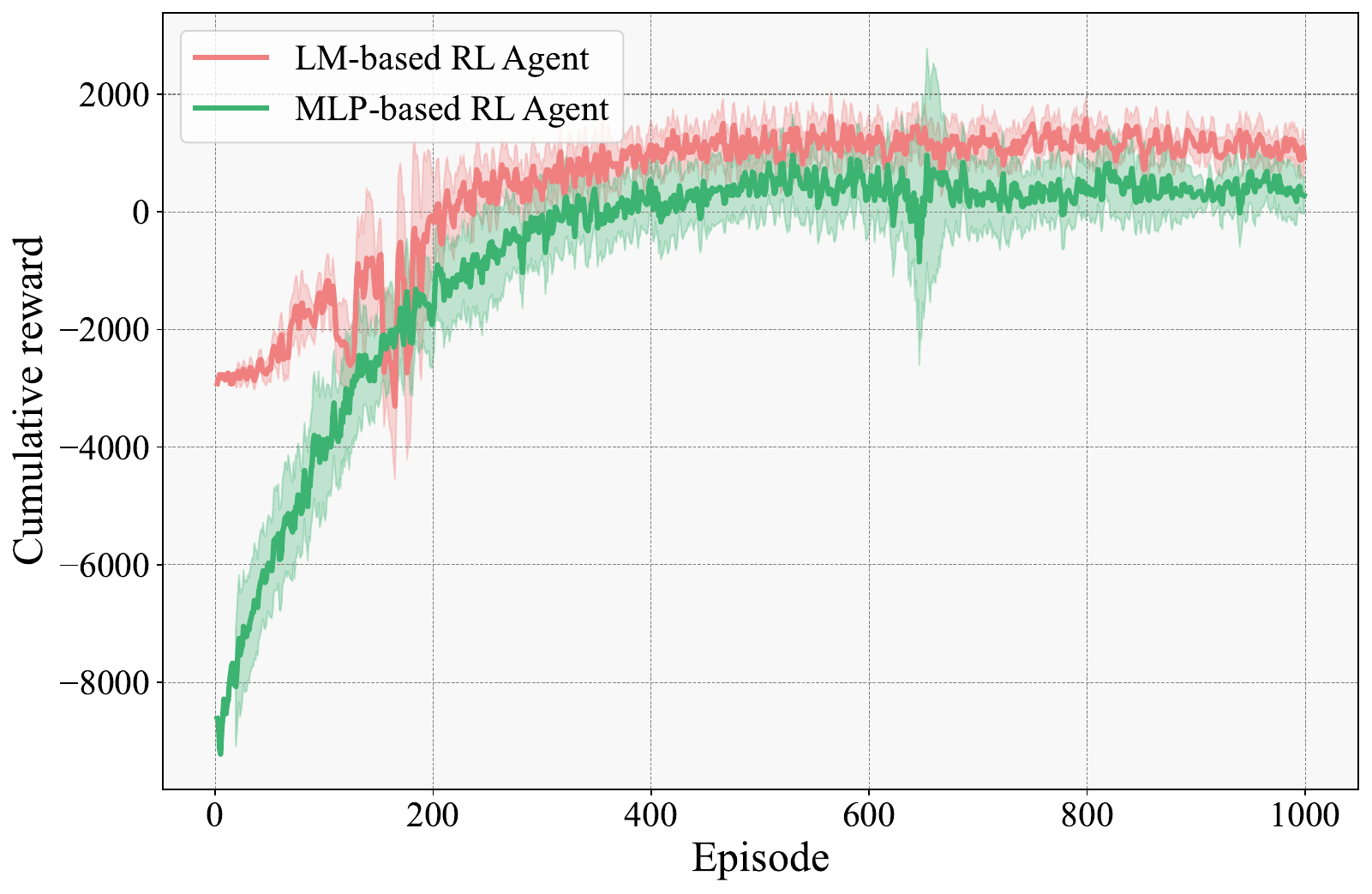}
   \caption{Cumulative reward versus episodes for policy training under RF1}
   \label{fig:reward}
   \vspace{-7mm}
\end{figure}

\subsection{Results of Experiments}
\label{exp: results}
The evaluation outcomes for the trained policies in both the Florida and Zaragoza case studies are detailed in Table~\ref{table:TP_results}, Table~\ref{table:TP_results2}, and Figure~\ref{fig:reward}. It's important to note that these results may not entirely reflect the optimal potential of the policies due to the random initialization of the Q-network and its episodic updates. Additionally, further refinement through hyperparameter tuning might yield more competitive outcomes. However, such tuning was intentionally avoided in this study to maintain a focus on generalizability and fair evaluation. Despite these deliverable-introduced constraints, the chosen policies still illustrate the effectiveness of the LM-based RL agent in enhancing crop management strategies. These policies also effectively demonstrate how different RFs can influence training outcomes.

The evaluation results, as detailed in Table \ref{table:TP_results} and Table \ref{table:TP_results2}, indicate that the proposed LM-based RL agent outperforms previous SoTA and empirical baselines in most metrics and scenarios. Notably, the LM-based RL agent consistently utilizes lower amounts of nitrogen and generally requires less irrigation, yet it manages to secure higher yields. These improvements are consistent across various reward functions  that prioritize different optimization objectives, underscoring the agent's adaptability and robustness in optimizing for diverse agricultural goals. The findings validate the previous hypothesis that language models have a heightened capacity to decipher complex crop management scenarios and simulate environments, ultimately leading to the discovery of more optimal management practices. Compared with the baseline policies, the RL-trained policies achieve a 49\% and a 67\% increase in terms of profit, i.e., RF1, and almost a 445\% and a 37\% increase in terms of RF5 for the Florida case and Zaragoza case, respectively. Notably, the enormous negative values of the cumulative rewards of Trained Policy 3 from both case studies are the results of the large amounts of N input, which are not punished during training with RF 3.

Consistent with prior studies \cite{tao2022optimizing}, the choice of reward function significantly influences the strategy of policies trained with LM-based RL agents. For instance, when trained with RF2, which posits irrigation water as a free resource, Trained Policy 2 tends to maximize irrigation while minimizing nitrogen input. This approach leads to the highest yield and cumulative reward as per the criteria of RF2. In contrast, RF3 assumes zero cost for nitrogen fertilizer, prompting Trained Policy 3 to favor high nitrogen use and minimal irrigation in both case studies. Under RF4, which reflects a doubled cost of nitrogen fertilizer in comparison to RF1, Trained Policy 4 leads to a reduction in nitrogen use. Despite the reduced nitrogen input, this policy still achieves a substantial yield and notably saves over 64\% of water resources, indicating the agent's capability to find a balance between cost efficiency and agricultural output.

In general, the results presented showcase the state-of-the-art capabilities of the LM-RL framework in optimizing crop management. This optimization is proven to be effective under various criteria, across different geographic locations, and within diverse environmental conditions. The framework's adaptability is highlighted by its ability to consistently apply LM-RL training to discover optimal management policies that align with specific targets, as dictated by the design of the chosen reward function. This flexibility and effectiveness affirm the potential of LM-RL as a powerful tool for agricultural management and decision-making.

\begin{table*}[ht]
\small
\centering
\begin{tabular}{lrrrrrrrr}
\toprule
Zaragoza Case & \makecell{N Input \\ (kg/ha) $\downarrow$} & \makecell{Irrigation\\ (L/m$^2$) $\downarrow$} & \makecell{Yield\\ (kg/ha) $\uparrow$} &RF1 $\uparrow$ &RF2 $\uparrow$ &RF3 $\uparrow$ &RF4 $\uparrow$ &RF5 $\uparrow$ \\ \midrule
Empirical Baseline         & 250     & 752            & 10990  &712                         &1539                        &909                        & 514                         & 1176
\\  \midrule
Policy1: Traditional Agent      & {240}     & {330}              & 10477  &{1103}                        &1466                       &1292                       &913                        & {1525}
\\ 
\rowcolor{lightgray}  Policy1: LM-based Agent (Ours)       & \textbf{160}     & {354}            & \textbf{10806}  &\textbf{1192}                        &\textbf{1581}                       &\textbf{1318}                       &\textbf{1065}                        & \textbf{1617} 
\\  \midrule

Policy2: Traditional Agent        & 200     & 1068             & 10923    &393                         & {1568}               &551                        &235                         &888                    \\ 
\rowcolor{lightgray}  Policy2: LM-based Agent (Ours)       & \textbf{160}     & \textbf{1032}             &\textbf{10856}           & \textbf{453}  &\textbf{1588}                        &\textbf{580}                       &\textbf{327}                                   & \textbf{964} 
\\  \midrule

Policy3: Traditional Agent       & 10640     & {324}               & 10626    &-7083                      &-6727                     & {1323}                      &{-1.5e4}                       &-8839                    \\
\rowcolor{lightgray}  Policy3: LM-based Agent (Ours)       & \textbf{10000}     & {342}            & \textbf{10903}                        &\textbf{-6553}                       &\textbf{-6177}                       &{1347 }         &\textbf{-8161}                 & \textbf{-8161} 
\\  \midrule

Policy4: Traditional Agent       & 120     & 336              & {9601}  &1053                        &1422                        &1147                        &{958}                        &1464     \\
\rowcolor{lightgray}  Policy4: LM-based Agent (Ours)       & \textbf{160}     & \textbf{348}            & \textbf{10250}  &\textbf{1110}                        &\textbf{1493}                       &\textbf{1268}                       &\textbf{984}                        & \textbf{1542} 
\\  \midrule

Policy5: Traditional Agent       & 200     & 390              & 10589 &1086              &1515                        &1244              &928               &1528     
\\  

\rowcolor{lightgray}  Policy5: LM-based Agent (Ours)       & \textbf{160}     & \textbf{362}            & \textbf{10660}  &\textbf{1160}                        &\textbf{1558}                       &\textbf{1286}                       &\textbf{1033}                        & \textbf{1610} 

\\ \bottomrule

\end{tabular}
\caption{The evaluation results of our trained policies, comparing them with previous SoTA methods and baseline policies. `Policy x' refers to the policy optimized using the reward function `RF x'. The `RF x' column details the cumulative rewards for each policy, calculated in accordance with `RF x'. Details of each reward function can be found in Table \ref{table:weight}. The best value is highlighted in \textbf{bold}.}
\label{table:TP_results2}
\vspace{-3mm}
\end{table*}

\subsection{Ablation Studies}
\label{exp: ablation}
\subsubsection{Training Separately on Fertilization and Irrigation}
In our previous research endeavors, we concurrently optimized N fertilization and irrigation practices, subsequently comparing these results against both established baseline practices and previous SoTAs. To further elucidate the efficacy of this joint optimization approach, this section introduces an ablation study wherein the management policies for N fertilization and irrigation were trained independently.  Specifically, while one practice was subject to optimization, the other adhered to established baseline methods. For instance, when optimizing an N management policy, the irrigation management followed the predefined baseline protocol, and vice versa. To be specific, experiments were conducted within the framework of the Florida case study, utilizing RF1 to guide the optimization process. The results, delineated in Table \ref{table:baseline}, provide a clear indication of the advantages inherent in the simultaneous optimization of N fertilization and irrigation management, as opposed to the independent optimization of each practice. This finding reveals that synergistically managing nitrogen fertilization and irrigation together yields superior agricultural outcomes compared to optimizing each practice in isolation.

\begin{table}[t]
\small
\centering
\begin{tabular}{l c c}\toprule
Fertilization & Irrigation &RF1$\uparrow$\\ \midrule
Baseline N Fertilization & Baseline Irrigation        & 984                \\ 
Baseline N Fertilization & Training Irrigation     &1376                       \\ 
Training N Fertilization & Baseline Irrigation       &1157                 \\ \hdashline
Training N Fertilization & Training Irrigation      &\textbf{1464}                 \\
\bottomrule
\end{tabular}
\caption{Performance comparison of the trained policies on both N fertilization and irrigation with the trained policies on either N fertilization or irrigation. The best values are shown in  \textbf{bold}.}
\label{table:baseline}
\vspace{-5mm}
\end{table}

\subsubsection{Exploration of Framework}
In order to investigate the most effective framework of RL agents, an ablation study was conducted. This study aimed to ascertain the impact of the framework's structure on management practices. Aligning with the setup of our previous experiments, we present the results for the Florida case using the reward function RF1. The outcomes, as depicted in Table \ref{table:adaptation}, indicate that employing a three-layer MLP yields the best results with a traditional RL agent. However, a notable decline in performance is observed when scaling the agent size from an MLP to a ResNet152 \cite{he2016deep}. This performance drop suggests the occurrence of overfitting within the RL framework, implying that simply increasing the size of the neural network does not necessarily enhance the exploration of optimal management practices.

Contrastingly, the use of LMs, such as Distilled Bert, demonstrated a different trend. Not only did the LM exhibit improved performance, but it also provided valuable insights. The results suggest that LMs possess a unique ability to comprehend the underlying patterns and logic of crop and environmental models. This capability enables them to pinpoint more optimal solutions while successfully circumventing the issue of overfitting, which was observed with larger neural network models.

\section{Path to Deployment}
\label{sec:deployment}
The effectiveness of management policies trained within the DSSAT-simulated environment may not directly translate to real-world scenarios. This potential discrepancy arises from uncertainties in weather conditions and differences between the crop models used for training and actual agricultural systems. This phenomenon, known as the \textit{sim-to-real gap} \cite{zhao2020sim}, highlights a common challenge in applying RL policies, developed and refined in simulated settings, to practical, real-world environments. 

\subsection{Closing the Sim-To-Real Gap}
To enhance the robustness of our trained management policies against the challenges posed by the \textit{sim-to-real gap}, we plan to incorporate \textit{domain and dynamics randomization} techniques, as suggested in previous studies \cite{tobin2017domainrand,peng2018sim2real}. This approach involves introducing variations in critical parameters of the model and randomizing weather conditions during policy training. Such perturbations are intended to ``force" the policies to become resilient to uncertainties in both the model and weather conditions.

While the primary focus of our current work is to establish the LM-based RL framework for crop management and to assess its effectiveness, we acknowledge the importance of addressing the robustness of these policies in real-world scenarios. Therefore, we aim to delve into this aspect in a forthcoming study, which will specifically target and evaluate the robustness of our LM-based RL policies against real-world variabilities and uncertainties.

\begin{table}[t]
\small
\centering
\begin{tabular}{l cc}
\hline
\toprule
Model Architecture & \# of Parameters & RF1$\uparrow$ \\ 
\midrule
Three-layer MLP & 0.2M & 1425 \\ 
Five-layer MLP & 0.5M & 1312 \\ \hdashline
ResNet18 & 11.0M & 510 \\ 
ResNet50 & 25.6M & 230 \\ 
ResNet101 & 44.7M & 107 \\ 
ResNet152 & 60.4M & 110 \\ \hdashline
Distilled Bert & 60.3M & \textbf{1464} \\
\bottomrule
\end{tabular}
\caption{Performance comparison of different frameworks as RL agents. The best values are shown in \textbf{bold}.}
\label{table:adaptation}
\vspace{-5mm}
\end{table}


\subsection{Policy Evaluation with Measurement Noises}
In order to assess the robustness of our method against random measurement noises, we conducted experiments following previous work \cite{tao2022optimizing}. In practical scenarios, farmers rely on weather forecasts and soil moisture measurements to make informed decisions. However, these data sources often contain inaccuracies due to forecast errors and sensor limitations. To simulate this real-world scenario, we tested LM-based RL under policy 1 from the Florida case study by introducing random measurement noises to key observable state variables each day in the simulation. These noise values were determined based on the real-world accuracy data of weather forecasts and commonly used soil moisture meters \cite{cobos2010does,floehr2010weather,zhang2018short,heinemann2006forecasting}. For each variable of added noise, the policy's performance was evaluated 400 times, with the average cumulative reward and standard deviation reported. The results, detailed in Table \ref{table:noise}, indicate that temperature and rainfall data inaccuracies have the most significant impact on policy performance, while other variables have minimal effects. Such an observation is consistent with previous research \cite{tao2022optimizing}. Notably, even with accumulated noise with multiple variables, the trained policy managed to achieve an average cumulative reward of 1248.8. While 15.3\% lower than the reward in a noise-free environment, it is still considerably higher than that of the baseline policy. These findings demonstrate that the policies trained using our method can yield relatively satisfactory and robust results compared to baseline approaches, even under real-world scenarios.

\begin{table}[t]

\centering
\small
\resizebox{1 \columnwidth}{!}{
\begin{tabular}{l r r r r}\toprule
\makecell{Variables} & Noises & RF 1 &STD &\makecell{Decrease\\ (\%)}   \\ \midrule
Empirical Baseline     & N/A           & 984.4   & N/A    &N/A    \\ 
No Noise               &N/A            &1463.9   &N/A     &N/A \\
Soil water content     &-+0.02         &1463.9   &0.0     &0.0   \\ 
Soil water content     &-+0.05         &1462.2   &1.9     &0.1 \\
Temperature            &-+1            &1443.7   &89.4    &1.3   \\
Temperature            &-+2            &1289.0   &361.0   &11.9 \\
Solar Radiation        &-+2\%          &1468.5   &0.7     &0 \\
Solar Radiation        &-+10\%         &1468.8   &7.6     &0 \\
Rain Fall              &90 \% Acc. &1416.5   &220.7   &3.2\\
Leaf Area Index        &-+10\%         &1457.1   &1.2     &0.4 \\
Leaf Area Index        &-+20\%         &1451.8   &5.8     &0.8 \\ 
\hdashline
Soil water content         &-+0.02           &    &      &  \\ 
+Temperature               &-+2              &    &      &  \\ 
+Solar Radiation           &-+2\%            &1248.8    &386.8      &  15.3\\ 
+ Rain Fall                &90 \% Acc.           &    &      &\\ 
+ Leaf Area Index          &-+20\%           &    &      &\\ 
\bottomrule
\end{tabular} 
}
\caption{Performance of the LM-based RL with Policy1 under measurement noises evaluated with RF1. The decrease (\%) is calculated with respect to RF1, where no noise was applied.}
\label{table:noise}

\vspace{-5mm}
\end{table}

\section{Conclusion}

In this paper, we address the crucial challenge of optimizing crop management to maximize yield while minimizing management costs and environmental impacts. We present an innovative framework that combines deep reinforcement learning, language models, and crop simulations using Gym-DSSAT. The experimental results clearly demonstrate that Language Model-based Reinforcement Learning agents surpass baseline models and significantly outperform existing  SoTA methods. This enhanced performance stems from the LM-RL agents' capacity to dynamically adjust their strategies according to different reward function designs, coupled with their ability to think and infer like expert agronomists. This dual capability enables them to maximize rewards in a variety of scenarios. Crucially, the framework has proven effective even in the presence of measurement noise in observable state variables, which is particularly promising for real-world applications.

We aspire for our work to serve as a proof of concept for the potential of LMs as adept agronomists, sparking interest and motivating further exploration in this area. The ultimate goal is to encourage researchers and practitioners to investigate and implement more advanced language models in practical agricultural settings. We believe that such advancements could significantly contribute to the evolution of agricultural technology, leading to smarter, more efficient, and sustainable farming practices worldwide.



{
    \small
    \bibliographystyle{ieeenat_fullname}
    \bibliography{main}

\begin{thebibliography}{84}
\providecommand{\natexlab}[1]{#1}
\providecommand{\url}[1]{\texttt{#1}}
\expandafter\ifx\csname urlstyle\endcsname\relax
  \providecommand{\doi}[1]{doi: #1}\else
  \providecommand{\doi}{doi: \begingroup \urlstyle{rm}\Url}\fi

\bibitem[Ahn et~al.(2022)Ahn, Brohan, Brown, Chebotar, Cortes, David, Finn, Fu, Gopalakrishnan, Hausman, et~al.]{ahn2022can}
Michael Ahn, Anthony Brohan, Noah Brown, Yevgen Chebotar, Omar Cortes, Byron David, Chelsea Finn, Chuyuan Fu, Keerthana Gopalakrishnan, Karol Hausman, et~al.
\newblock Do as i can, not as i say: Grounding language in robotic affordances.
\newblock \emph{arXiv preprint arXiv:2204.01691}, 2022.

\bibitem[Ara et~al.(2021)Ara, Turner, Harrison, Monjardino, DeVoil, and Rodriguez]{ara2021application}
Iffat Ara, Lydia Turner, Matthew~Tom Harrison, Marta Monjardino, Peter DeVoil, and Daniel Rodriguez.
\newblock Application, adoption and opportunities for improving decision support systems in irrigated agriculture: A review.
\newblock \emph{Agricultural Water Management}, 257:\penalty0 107161, 2021.

\bibitem[Ashcraft and Karra(2021)]{ashcraft2021ml-aided}
Chace Ashcraft and Kiran Karra.
\newblock Machine learning aided crop yield optimization.
\newblock \emph{arXiv preprint arXiv:2111.00963}, 2021.

\bibitem[Asseng et~al.(2013)Asseng, Ewert, Rosenzweig, Jones, Hatfield, Ruane, Boote, Thorburn, R{\"o}tter, Cammarano, et~al.]{asseng2013uncertainty}
Senthold Asseng, Frank Ewert, Cynthia Rosenzweig, James~W Jones, Jerry~L Hatfield, Alex~C Ruane, Kenneth~J Boote, Peter~J Thorburn, Reimund~P R{\"o}tter, Davide Cammarano, et~al.
\newblock Uncertainty in simulating wheat yields under climate change.
\newblock \emph{Nature climate change}, 3\penalty0 (9):\penalty0 827--832, 2013.

\bibitem[Attonaty et~al.(1997)Attonaty, Chatelin, Garcia, and Ndiaye]{attonaty1997using}
J-M Attonaty, M-H Chatelin, F Garcia, and S Ndiaye.
\newblock Using extended machine learning and simulation technics to design crop management strategies.
\newblock In \emph{EFITA First European Conference for Information Technology in Agriculture, Copenhagen (Denmark)}, 1997.

\bibitem[Bassu et~al.(2014)Bassu, Brisson, Durand, Boote, Lizaso, Jones, Rosenzweig, Ruane, Adam, Baron, et~al.]{bassu2014various}
Simona Bassu, Nadine Brisson, Jean-Louis Durand, Kenneth Boote, Jon Lizaso, James~W Jones, Cynthia Rosenzweig, Alex~C Ruane, Myriam Adam, Christian Baron, et~al.
\newblock How do various maize crop models vary in their responses to climate change factors?
\newblock \emph{Global change biology}, 20\penalty0 (7):\penalty0 2301--2320, 2014.

\bibitem[Brockman et~al.(2016)Brockman, Cheung, Pettersson, Schneider, Schulman, Tang, and Zaremba]{brockman2016openai}
Greg Brockman, Vicki Cheung, Ludwig Pettersson, Jonas Schneider, John Schulman, Jie Tang, and Wojciech Zaremba.
\newblock Openai gym.
\newblock \emph{arXiv preprint arXiv:1606.01540}, 2016.

\bibitem[Chen et~al.(2023{\natexlab{a}})Chen, Xia, Ichter, Rao, Gopalakrishnan, Ryoo, Stone, and Kappler]{chen2023open}
Boyuan Chen, Fei Xia, Brian Ichter, Kanishka Rao, Keerthana Gopalakrishnan, Michael~S Ryoo, Austin Stone, and Daniel Kappler.
\newblock Open-vocabulary queryable scene representations for real world planning.
\newblock In \emph{2023 IEEE International Conference on Robotics and Automation (ICRA)}, pages 11509--11522. IEEE, 2023{\natexlab{a}}.

\bibitem[Chen and Mueller(2023)]{chen2023quantifying}
Jiuhai Chen and Jonas Mueller.
\newblock Quantifying uncertainty in answers from any language model via intrinsic and extrinsic confidence assessment.
\newblock \emph{arXiv preprint arXiv:2308.16175}, 2023.

\bibitem[Chen et~al.(2023{\natexlab{b}})Chen, Chen, Huang, and Zhou]{chen2023you}
Jiuhai Chen, Lichang Chen, Heng Huang, and Tianyi Zhou.
\newblock When do you need chain-of-thought prompting for chatgpt?
\newblock \emph{arXiv preprint arXiv:2304.03262}, 2023{\natexlab{b}}.

\bibitem[Chen et~al.(2017{\natexlab{a}})Chen, Kearns, Fozard, and Li]{chen2017personalized}
Suiyao Chen, William~D Kearns, James~L Fozard, and Mingyang Li.
\newblock Personalized fall risk assessment for long-term care services improvement.
\newblock In \emph{2017 Annual Reliability and Maintainability Symposium (RAMS)}, pages 1--7. IEEE, 2017{\natexlab{a}}.

\bibitem[Chen et~al.(2017{\natexlab{b}})Chen, Lu, and Li]{chen2017multi}
Suiyao Chen, Lu Lu, and Mingyang Li.
\newblock Multi-state reliability demonstration tests.
\newblock \emph{Quality Engineering}, 29\penalty0 (3):\penalty0 431--445, 2017{\natexlab{b}}.

\bibitem[Chen et~al.(2018)Chen, Lu, Xiang, Lu, and Li]{chen2018data}
Suiyao Chen, Lu Lu, Yisha Xiang, Qing Lu, and Mingyang Li.
\newblock A data heterogeneity modeling and quantification approach for field pre-assessment of chloride-induced corrosion in aging infrastructures.
\newblock \emph{Reliability Engineering \& System Safety}, 171:\penalty0 123--135, 2018.

\bibitem[Chen et~al.(2019)Chen, Kong, Sun, Meng, and Li]{chen2019claims}
Suiyao Chen, Nan Kong, Xuxue Sun, Hongdao Meng, and Mingyang Li.
\newblock Claims data-driven modeling of hospital time-to-readmission risk with latent heterogeneity.
\newblock \emph{Health care management science}, 22:\penalty0 156--179, 2019.

\bibitem[Chen et~al.(2020)Chen, Lu, Zhang, and Li]{chen2020optimal}
Suiyao Chen, Lu Lu, Qiong Zhang, and Mingyang Li.
\newblock Optimal binomial reliability demonstration tests design under acceptance decision uncertainty.
\newblock \emph{Quality Engineering}, 32\penalty0 (3):\penalty0 492--508, 2020.

\bibitem[Chen et~al.(2023{\natexlab{c}})Chen, Wu, Hovakimyan, and Yao]{chen2023recontab}
Suiyao Chen, Jing Wu, Naira Hovakimyan, and Handong Yao.
\newblock Recontab: Regularized contrastive representation learning for tabular data.
\newblock \emph{arXiv preprint arXiv:2310.18541}, 2023{\natexlab{c}}.

\bibitem[Cheng et~al.(2022)Cheng, Zhao, Wang, Block, and Hovakimyan]{cheng2022improving}
Yikun Cheng, Pan Zhao, Fanxin Wang, Daniel~J Block, and Naira Hovakimyan.
\newblock Improving the robustness of reinforcement learning policies with $\mathcal{L}_1$ adaptive control.
\newblock \emph{IEEE Robotics and Automation Letters}, 7\penalty0 (3):\penalty0 6574--6581, 2022.

\bibitem[Cheng et~al.(2023)Cheng, Zhao, and Hovakimyan]{cheng2023safe}
Yikun Cheng, Pan Zhao, and Naira Hovakimyan.
\newblock Safe and efficient reinforcement learning using disturbance-observer-based control barrier functions.
\newblock In \emph{Learning for Dynamics and Control Conference}, pages 104--115. PMLR, 2023.

\bibitem[Cobos and Devices(2010)]{cobos2010does}
Douglas~R Cobos and Decagon Devices.
\newblock Why does my soil moisture sensor read negative?
\newblock \emph{Online]. Available: http://manuals. decagon. com/RetiredandDiscontinued/Slicksandcontent/ Presentations/SoilMoisture301.pdf}, 2010.

\bibitem[Di and Cameron(2002)]{di2002nitrate}
HJ Di and KC Cameron.
\newblock Nitrate leaching in temperate agroecosystems: sources, factors and mitigating strategies.
\newblock \emph{Nutrient cycling in agroecosystems}, 64:\penalty0 237--256, 2002.

\bibitem[Ding and Wong(2019)]{ding2019confidence}
Zhicheng Ding and Edward Wong.
\newblock Confidence trigger detection: an approach to build real-time tracking-by-detection system.
\newblock \emph{arXiv preprint arXiv:1902.00615}, 2019.

\bibitem[Floehr(2010)]{floehr2010weather}
Eric Floehr.
\newblock Weather forecast accuracy analysis.
\newblock In \emph{Proc of the 9th Python in Science Conference, SciPy}, pages 36--39, 2010.

\bibitem[Garcia(1999)]{garcia1999rl-wheat}
Fr{\'e}d{\'e}rick Garcia.
\newblock Use of reinforcement learning and simulation to optimize wheat crop technical management.
\newblock In \emph{Proceedings of the International Congress on Modelling and Simulation}, pages 801--806, 1999.

\bibitem[He et~al.(2016)He, Zhang, Ren, and Sun]{he2016deep}
Kaiming He, Xiangyu Zhang, Shaoqing Ren, and Jian Sun.
\newblock Deep residual learning for image recognition.
\newblock In \emph{Proceedings of the IEEE conference on computer vision and pattern recognition}, pages 770--778, 2016.

\bibitem[Heinemann et~al.(2006)Heinemann, Lorenz, and Girodo]{heinemann2006forecasting}
Detlev Heinemann, Elke Lorenz, and Marco Girodo.
\newblock Forecasting of solar radiation.
\newblock \emph{Solar energy resource management for electricity generation from local level to global scale. Nova Science Publishers, New York}, pages 83--94, 2006.

\bibitem[Hu et~al.(2023)Hu, Zhang, Yu, Zhuang, and Xiong]{hu2023many}
Zhengyu Hu, Jieyu Zhang, Yue Yu, Yuchen Zhuang, and Hui Xiong.
\newblock How many validation labels do you need? exploring the design space of label-efficient model ranking.
\newblock \emph{arXiv preprint arXiv:2312.01619}, 2023.

\bibitem[Huang et~al.(2022{\natexlab{a}})Huang, Abbeel, Pathak, and Mordatch]{huang2022language}
Wenlong Huang, Pieter Abbeel, Deepak Pathak, and Igor Mordatch.
\newblock Language models as zero-shot planners: Extracting actionable knowledge for embodied agents.
\newblock In \emph{International Conference on Machine Learning}, pages 9118--9147. PMLR, 2022{\natexlab{a}}.

\bibitem[Huang et~al.(2022{\natexlab{b}})Huang, Xia, Xiao, Chan, Liang, Florence, Zeng, Tompson, Mordatch, Chebotar, et~al.]{huang2022inner}
Wenlong Huang, Fei Xia, Ted Xiao, Harris Chan, Jacky Liang, Pete Florence, Andy Zeng, Jonathan Tompson, Igor Mordatch, Yevgen Chebotar, et~al.
\newblock Inner monologue: Embodied reasoning through planning with language models.
\newblock \emph{arXiv preprint arXiv:2207.05608}, 2022{\natexlab{b}}.

\bibitem[Jones et~al.(2003)Jones, Hoogenboom, Porter, Boote, Batchelor, Hunt, Wilkens, Singh, Gijsman, and Ritchie]{jones2003dssat}
James~W Jones, Gerrit Hoogenboom, Cheryl~H Porter, Ken~J Boote, William~D Batchelor, LA Hunt, Paul~W Wilkens, Upendra Singh, Arjan~J Gijsman, and Joe~T Ritchie.
\newblock The {DSSAT} cropping system model.
\newblock \emph{European Journal of Agronomy}, 18\penalty0 (3-4):\penalty0 235--265, 2003.

\bibitem[Jones et~al.(2017)Jones, Antle, Basso, Boote, Conant, Foster, Godfray, Herrero, Howitt, Janssen, et~al.]{jones2017brief}
James~W Jones, John~M Antle, Bruno Basso, Kenneth~J Boote, Richard~T Conant, Ian Foster, H~Charles~J Godfray, Mario Herrero, Richard~E Howitt, Sander Janssen, et~al.
\newblock Brief history of agricultural systems modeling.
\newblock \emph{Agricultural systems}, 155:\penalty0 240--254, 2017.

\bibitem[Kallenberg et~al.(2023)Kallenberg, Overweg, van Bree, and Athanasiadis]{kallenberg2023nitrogen}
Michiel~GJ Kallenberg, Hiske Overweg, Ron van Bree, and Ioannis~N Athanasiadis.
\newblock Nitrogen management with reinforcement learning and crop growth models.
\newblock \emph{Environmental Data Science}, 2:\penalty0 e34, 2023.

\bibitem[Kaufmann et~al.(2018)Kaufmann, Loquercio, Ranftl, Dosovitskiy, Koltun, and Scaramuzza]{kaufmann2018droneracing-rl}
Elia Kaufmann, Antonio Loquercio, Rene Ranftl, Alexey Dosovitskiy, Vladlen Koltun, and Davide Scaramuzza.
\newblock Deep drone racing: {Learning agile flight in dynamic environments}.
\newblock In \emph{Conference on Robot Learning}, pages 133--145, 2018.

\bibitem[Kim et~al.(2024)Kim, Baldi, and McAleer]{kim2024language}
Geunwoo Kim, Pierre Baldi, and Stephen McAleer.
\newblock Language models can solve computer tasks.
\newblock \emph{Advances in Neural Information Processing Systems}, 36, 2024.

\bibitem[Kingma and Ba(2014)]{kingma2014adam}
Diederik~P Kingma and Jimmy Ba.
\newblock Adam: A method for stochastic optimization.
\newblock \emph{arXiv preprint arXiv:1412.6980}, 2014.

\bibitem[Lai et~al.(2024)Lai, Zhang, and Chen]{lai2024adaptive}
Zhixin Lai, Xuesheng Zhang, and Suiyao Chen.
\newblock Adaptive ensembles of fine-tuned transformers for llm-generated text detection.
\newblock \emph{arXiv preprint arXiv:2403.13335}, 2024.

\bibitem[Li et~al.(2022)Li, You, Funakoshi, and Okumura]{li2022tip}
Dongyuan Li, Jingyi You, Kotaro Funakoshi, and Manabu Okumura.
\newblock A-tip: attribute-aware text infilling via pre-trained language model.
\newblock In \emph{Proceedings of the 29th International Conference on Computational Linguistics}, pages 5857--5869, 2022.

\bibitem[Li et~al.(2023{\natexlab{a}})Li, Wang, Funakoshi, and Okumura]{li2023joyful}
Dongyuan Li, Yusong Wang, Kotaro Funakoshi, and Manabu Okumura.
\newblock Joyful: Joint modality fusion and graph contrastive learning for multimodal emotion recognition.
\newblock \emph{arXiv preprint arXiv:2311.11009}, 2023{\natexlab{a}}.

\bibitem[Li et~al.(2023{\natexlab{b}})Li, Zhu, Lu, and Yin]{li-etal-2023-synthetic}
Zhuoyan Li, Hangxiao Zhu, Zhuoran Lu, and Ming Yin.
\newblock Synthetic data generation with large language models for text classification: Potential and limitations.
\newblock In \emph{Proceedings of the 2023 Conference on Empirical Methods in Natural Language Processing}, pages 10443--10461, Singapore, 2023{\natexlab{b}}. Association for Computational Linguistics.

\bibitem[Liang et~al.(2023)Liang, Huang, Xia, Xu, Hausman, Ichter, Florence, and Zeng]{liang2023code}
Jacky Liang, Wenlong Huang, Fei Xia, Peng Xu, Karol Hausman, Brian Ichter, Pete Florence, and Andy Zeng.
\newblock Code as policies: Language model programs for embodied control.
\newblock In \emph{2023 IEEE International Conference on Robotics and Automation (ICRA)}, pages 9493--9500. IEEE, 2023.

\bibitem[Liu and Zhu(2022)]{liu2022distributed}
Shicheng Liu and Minghui Zhu.
\newblock Distributed inverse constrained reinforcement learning for multi-agent systems.
\newblock \emph{Advances in Neural Information Processing Systems}, 35:\penalty0 33444--33456, 2022.

\bibitem[Liu and Zhu(2023)]{liu2023meta}
Shicheng Liu and Minghui Zhu.
\newblock Meta inverse constrained reinforcement learning: Convergence guarantee and generalization analysis.
\newblock In \emph{The Twelfth International Conference on Learning Representations}, 2023.

\bibitem[Liu and Zhu(2024)]{liu2024learning}
Shicheng Liu and Minghui Zhu.
\newblock Learning multi-agent behaviors from distributed and streaming demonstrations.
\newblock \emph{Advances in Neural Information Processing Systems}, 36, 2024.

\bibitem[Liu et~al.(2024{\natexlab{a}})Liu, {\v{S}}krjanec, and Demberg]{liu2024temperature}
Tong Liu, Iza {\v{S}}krjanec, and Vera Demberg.
\newblock Temperature-scaling surprisal estimates improve fit to human reading times--but does it do so for the “right reasons”?
\newblock In \emph{ICLR 2024 Workshop on Representational Alignment}, 2024{\natexlab{a}}.

\bibitem[Liu et~al.(2024{\natexlab{b}})Liu, Xu, Qiao, Jiang, and Chen]{liu2024news}
Tianrui Liu, Changxin Xu, Yuxin Qiao, Chufeng Jiang, and Weisheng Chen.
\newblock News recommendation with attention mechanism.
\newblock \emph{Journal of Industrial Engineering and Applied Science}, 2\penalty0 (1):\penalty0 21--26, 2024{\natexlab{b}}.

\bibitem[Liu et~al.(2024{\natexlab{c}})Liu, Xu, Qiao, Jiang, and Yu]{liu2024particle}
Tianrui Liu, Changxin Xu, Yuxin Qiao, Chufeng Jiang, and Jiqiang Yu.
\newblock Particle filter slam for vehicle localization.
\newblock \emph{Journal of Industrial Engineering and Applied Science}, 2\penalty0 (1):\penalty0 27--31, 2024{\natexlab{c}}.

\bibitem[Madondo et~al.(2023)Madondo, Azmat, Dipietro, Horesh, Jacobs, Bawa, Srinivasan, and O'Donncha]{madondo2023swat}
Malvern Madondo, Muneeza Azmat, Kelsey Dipietro, Raya Horesh, Michael Jacobs, Arun Bawa, Raghavan Srinivasan, and Fearghal O'Donncha.
\newblock A swat-based reinforcement learning framework for crop management.
\newblock \emph{arXiv preprint arXiv:2302.04988}, 2023.

\bibitem[Malik et~al.(2019)Malik, Isla, and Dechmi]{malik2019dssat}
Wafa Malik, Ramon Isla, and Farida Dechmi.
\newblock Dssat-ceres-maize modelling to improve irrigation and nitrogen management practices under mediterranean conditions.
\newblock \emph{Agricultural Water Management}, 213:\penalty0 298--308, 2019.

\bibitem[Mandrini et~al.(2022)Mandrini, Pittelkow, Archontoulis, Kanter, and Martin]{mandrini2022exploring}
German Mandrini, Cameron~Mark Pittelkow, Sotirios Archontoulis, David Kanter, and Nicolas~F Martin.
\newblock Exploring trade-offs between profit, yield, and the environmental footprint of potential nitrogen fertilizer regulations in the us midwest.
\newblock \emph{Frontiers in plant science}, 13, 2022.

\bibitem[Mees et~al.(2023)Mees, Borja-Diaz, and Burgard]{mees2023grounding}
Oier Mees, Jessica Borja-Diaz, and Wolfram Burgard.
\newblock Grounding language with visual affordances over unstructured data.
\newblock In \emph{2023 IEEE International Conference on Robotics and Automation (ICRA)}, pages 11576--11582. IEEE, 2023.

\bibitem[Mnih et~al.(2015)Mnih, Kavukcuoglu, Silver, et~al.]{mnih2015human-deepRL}
Volodymyr Mnih, Koray Kavukcuoglu, David Silver, et~al.
\newblock Human-level control through deep reinforcement learning.
\newblock \emph{Nature}, 518\penalty0 (7540):\penalty0 529--533, 2015.

\bibitem[Overweg et~al.(2021)Overweg, Berghuijs, and Athanasiadis]{overweg2021cropgym}
Hiske Overweg, Herman~NC Berghuijs, and Ioannis~N Athanasiadis.
\newblock Cropgym: a reinforcement learning environment for crop management.
\newblock \emph{arXiv preprint arXiv:2104.04326}, 2021.

\bibitem[Palmer et~al.(2013)Palmer, Cooper, T{\'e}tard-Jones, et~al.]{palmer2013influence}
Mike~W Palmer, Julia Cooper, Catherine T{\'e}tard-Jones, et~al.
\newblock The influence of organic and conventional fertilisation and crop protection practices, preceding crop, harvest year and weather conditions on yield and quality of potato ({Solanum} tuberosum) in a long-term management trial.
\newblock \emph{European Journal of Agronomy}, 49:\penalty0 83--92, 2013.

\bibitem[Park et~al.(2023)Park, O'Brien, Cai, Morris, Liang, and Bernstein]{park2023generative}
Joon~Sung Park, Joseph O'Brien, Carrie~Jun Cai, Meredith~Ringel Morris, Percy Liang, and Michael~S Bernstein.
\newblock Generative agents: Interactive simulacra of human behavior.
\newblock In \emph{Proceedings of the 36th Annual ACM Symposium on User Interface Software and Technology}, pages 1--22, 2023.

\bibitem[Peng et~al.(2018)Peng, Andrychowicz, Zaremba, and Abbeel]{peng2018sim2real}
Xue~Bin Peng, Marcin Andrychowicz, Wojciech Zaremba, and Pieter Abbeel.
\newblock Sim-to-real transfer of robotic control with dynamics randomization.
\newblock In \emph{IEEE International Conference on Robotics and Automation (ICRA)}, pages 3803--3810, 2018.

\bibitem[Raman et~al.(2022)Raman, Cohen, Rosen, Idrees, Paulius, and Tellex]{raman2022planning}
Shreyas~Sundara Raman, Vanya Cohen, Eric Rosen, Ifrah Idrees, David Paulius, and Stefanie Tellex.
\newblock Planning with large language models via corrective re-prompting.
\newblock In \emph{NeurIPS 2022 Foundation Models for Decision Making Workshop}, 2022.

\bibitem[Reddy et~al.(2003)Reddy, Reddy, Bidinger, and Bl{\"u}mmel]{reddy2003crop}
BVS Reddy, P~Sanjana Reddy, F Bidinger, and Michael Bl{\"u}mmel.
\newblock Crop management factors influencing yield and quality of crop residues.
\newblock \emph{Field Crops Research}, 84\penalty0 (1-2):\penalty0 57--77, 2003.

\bibitem[Romain et~al.(2022)Romain, Philippe, Julien, Odalric-Ambrym, David, et~al.]{romain2022gym}
Gautron Romain, Preux Philippe, Bigot Julien, Maillard Odalric-Ambrym, Emukpere David, et~al.
\newblock gym-dssat: a crop model turned into a reinforcement learning environment.
\newblock \emph{arXiv preprint arXiv:2207.03270}, 2022.

\bibitem[Sanh et~al.(2019)Sanh, Debut, Chaumond, and Wolf]{sanh2019distilbert}
Victor Sanh, Lysandre Debut, Julien Chaumond, and Thomas Wolf.
\newblock Distilbert, a distilled version of bert: smaller, faster, cheaper and lighter.
\newblock \emph{arXiv preprint arXiv:1910.01108}, 2019.

\bibitem[Searchinger et~al.(2019)Searchinger, Waite, Hanson, Ranganathan, Dumas, Matthews, and Klirs]{searchinger2019creating}
Tim Searchinger, Richard Waite, Craig Hanson, Janet Ranganathan, Patrice Dumas, Emily Matthews, and Carni Klirs.
\newblock Creating a sustainable food future: A menu of solutions to feed nearly 10 billion people by 2050. final report.
\newblock \emph{World Resources Institude}, 2019.

\bibitem[Shi et~al.(2017)Shi, Ling, Xue, Qin, Li, Lai, and Yang]{shi2017combining}
Ji-Ying Shi, Le-Tao Ling, Fei Xue, Zi-Jian Qin, Ya-Jing Li, Zhi-Xin Lai, and Ting Yang.
\newblock Combining incremental conductance and firefly algorithm for tracking the global mpp of pv arrays.
\newblock \emph{Journal of Renewable and Sustainable Energy}, 9\penalty0 (2), 2017.

\bibitem[Shinn et~al.(2024)Shinn, Cassano, Gopinath, Narasimhan, and Yao]{shinn2024reflexion}
Noah Shinn, Federico Cassano, Ashwin Gopinath, Karthik Narasimhan, and Shunyu Yao.
\newblock Reflexion: Language agents with verbal reinforcement learning.
\newblock \emph{Advances in Neural Information Processing Systems}, 36, 2024.

\bibitem[Skhiri and Dechmi(2012)]{skhiri2012impact}
Ahmed Skhiri and Farida Dechmi.
\newblock Impact of sprinkler irrigation management on the del reguero river (spain). i: Water balance and irrigation performance.
\newblock \emph{Agricultural Water Management}, 103:\penalty0 120--129, 2012.

\bibitem[Su et~al.(2024)Su, Jiang, Jin, Qiao, Xiao, Ma, Wei, Jing, Xu, and Lin]{su2024large}
Jing Su, Chufeng Jiang, Xin Jin, Yuxin Qiao, Tingsong Xiao, Hongda Ma, Rong Wei, Zhi Jing, Jiajun Xu, and Junhong Lin.
\newblock Large language models for forecasting and anomaly detection: A systematic literature review.
\newblock \emph{arXiv preprint arXiv:2402.10350}, 2024.

\bibitem[Sun et~al.(2017)Sun, Yang, Hu, Porter, Marek, and Hillyer]{sun2017rl-irrigation}
Lijia Sun, Yanxiang Yang, Jiang Hu, Dana Porter, Thomas Marek, and Charles Hillyer.
\newblock Reinforcement learning control for water-efficient agricultural irrigation.
\newblock In \emph{2017 IEEE International Symposium on Parallel and Distributed Processing with Applications and 2017 IEEE International Conference on Ubiquitous Computing and Communications (ISPA/IUCC)}, pages 1334--1341, 2017.

\bibitem[Tao et~al.(2022)Tao, Zhao, Wu, Martin, Harrison, Ferreira, Kalantari, and Hovakimyan]{tao2022optimizing}
Ran Tao, Pan Zhao, Jing Wu, Nicolas~F Martin, Matthew~T Harrison, Carla Ferreira, Zahra Kalantari, and Naira Hovakimyan.
\newblock Optimizing crop management with reinforcement learning and imitation learning.
\newblock \emph{arXiv preprint arXiv:2209.09991}, 2022.

\bibitem[Tobin et~al.(2017)Tobin, Fong, Ray, Schneider, Zaremba, and Abbeel]{tobin2017domainrand}
Josh Tobin, Rachel Fong, Alex Ray, Jonas Schneider, Wojciech Zaremba, and Pieter Abbeel.
\newblock Domain randomization for transferring deep neural networks from simulation to the real world.
\newblock In \emph{IEEE/RSJ International Conference on Intelligent Robots and Systems (IROS)}, pages 23--30, 2017.

\bibitem[Wang et~al.(2023{\natexlab{a}})Wang, Lu, Chen, and Li]{bingjie2023optimal}
Bingjie Wang, Lu Lu, Suiyao Chen, and Mingyang Li.
\newblock Optimal test design for reliability demonstration under multi-stage acceptance uncertainties.
\newblock \emph{Quality Engineering}, 0\penalty0 (0):\penalty0 1--14, 2023{\natexlab{a}}.

\bibitem[Wang et~al.(2023{\natexlab{b}})Wang, Li, Funakoshi, and Okumura]{wang2023emp}
Yusong Wang, Dongyuan Li, Kotaro Funakoshi, and Manabu Okumura.
\newblock Emp: emotion-guided multi-modal fusion and contrastive learning for personality traits recognition.
\newblock In \emph{Proceedings of the 2023 ACM International Conference on Multimedia Retrieval}, pages 243--252, 2023{\natexlab{b}}.

\bibitem[Wang et~al.(2023{\natexlab{c}})Wang, Wu, Hovakimyan, and Sun]{wang2023balanced}
Yite Wang, Jing Wu, Naira Hovakimyan, and Ruoyu Sun.
\newblock Balanced training for sparse gans.
\newblock In \emph{Thirty-seventh Conference on Neural Information Processing Systems}, 2023{\natexlab{c}}.

\bibitem[Wei et~al.(2022)Wei, Wang, Schuurmans, Bosma, Xia, Chi, Le, Zhou, et~al.]{wei2022chain}
Jason Wei, Xuezhi Wang, Dale Schuurmans, Maarten Bosma, Fei Xia, Ed Chi, Quoc~V Le, Denny Zhou, et~al.
\newblock Chain-of-thought prompting elicits reasoning in large language models.
\newblock \emph{Advances in Neural Information Processing Systems}, 35:\penalty0 24824--24837, 2022.

\bibitem[Wright et~al.(2022)Wright, Small, Mackowiak, Grabau, Devkota, and Paula-Moraes]{wright2022field}
David Wright, Ian Small, Cheryl Mackowiak, Zane Grabau, Pratap Devkota, and Silvana Paula-Moraes.
\newblock Field corn production guide: Ss-agr-85/ag202, rev. 8/2022.
\newblock \emph{EDIS}, 2022\penalty0 (4), 2022.

\bibitem[Wu et~al.(2022)Wu, Tao, Zhao, Martin, and Hovakimyan]{wu2022optimizing}
Jing Wu, Ran Tao, Pan Zhao, Nicolas~F Martin, and Naira Hovakimyan.
\newblock Optimizing nitrogen management with deep reinforcement learning and crop simulations.
\newblock In \emph{Proceedings of the IEEE/CVF Conference on Computer Vision and Pattern Recognition Workshops}, pages 1712--1720, 2022.

\bibitem[Wu et~al.(2023{\natexlab{a}})Wu, Hobbs, and Hovakimyan]{wu2023hallucination}
Jing Wu, Jennifer Hobbs, and Naira Hovakimyan.
\newblock Hallucination improves the performance of unsupervised visual representation learning.
\newblock In \emph{Proceedings of the IEEE/CVF International Conference on Computer Vision}, pages 16132--16143, 2023{\natexlab{a}}.

\bibitem[Wu et~al.(2023{\natexlab{b}})Wu, Hovakimyan, and Hobbs]{wu2023genco}
Jing Wu, Naira Hovakimyan, and Jennifer Hobbs.
\newblock Genco: An auxiliary generator from contrastive learning for enhanced few-shot learning in remote sensing.
\newblock \emph{arXiv preprint arXiv:2307.14612}, 2023{\natexlab{b}}.

\bibitem[Wu et~al.(2023{\natexlab{c}})Wu, Pichler, Marley, Wilson, Hovakimyan, and Hobbs]{wu2023extended}
Jing Wu, David Pichler, Daniel Marley, David Wilson, Naira Hovakimyan, and Jennifer Hobbs.
\newblock Extended agriculture-vision: An extension of a large aerial image dataset for agricultural pattern analysis.
\newblock \emph{arXiv preprint arXiv:2303.02460}, 2023{\natexlab{c}}.

\bibitem[Wu et~al.(2024)Wu, Chen, Zhao, Sergazinov, Li, Liu, Zhao, Xie, Guo, Ji, et~al.]{wu2024switchtab}
Jing Wu, Suiyao Chen, Qi Zhao, Renat Sergazinov, Chen Li, Shengjie Liu, Chongchao Zhao, Tianpei Xie, Hanqing Guo, Cheng Ji, et~al.
\newblock Switchtab: Switched autoencoders are effective tabular learners.
\newblock \emph{arXiv preprint arXiv:2401.02013}, 2024.

\bibitem[Xin et~al.(2023)Xin, Du, Wang, Yan, and Ding]{xin2023mmap}
Yi Xin, Junlong Du, Qiang Wang, Ke Yan, and Shouhong Ding.
\newblock Mmap: Multi-modal alignment prompt for cross-domain multi-task learning.
\newblock \emph{arXiv preprint arXiv:2312.08636}, 2023.

\bibitem[Yao et~al.(2022)Yao, Zhao, Yu, Du, Shafran, Narasimhan, and Cao]{yao2022react}
Shunyu Yao, Jeffrey Zhao, Dian Yu, Nan Du, Izhak Shafran, Karthik Narasimhan, and Yuan Cao.
\newblock React: Synergizing reasoning and acting in language models.
\newblock \emph{arXiv preprint arXiv:2210.03629}, 2022.

\bibitem[Zhang et~al.(2024)Zhang, Wang, Hu, Koh, and Ratner]{zhang2024trade}
Jieyu Zhang, Bohan Wang, Zhengyu Hu, Pang Wei~W Koh, and Alexander~J Ratner.
\newblock On the trade-off of intra-/inter-class diversity for supervised pre-training.
\newblock \emph{Advances in Neural Information Processing Systems}, 36, 2024.

\bibitem[Zhang et~al.(2018)Zhang, Jia, Gao, Song, and Leung]{zhang2018short}
Pengcheng Zhang, Yangyang Jia, Jerry Gao, Wei Song, and Hareton Leung.
\newblock Short-term rainfall forecasting using multi-layer perceptron.
\newblock \emph{IEEE Transactions on Big Data}, 6\penalty0 (1):\penalty0 93--106, 2018.

\bibitem[Zhao et~al.(2019)Zhao, Liu, Xiao, Hoogenboom, Boote, Kassie, Pavan, Shelia, Kim, Hernandez-Ochoa, et~al.]{zhao2019simple}
Chuang Zhao, Bing Liu, Liujun Xiao, Gerrit Hoogenboom, Kenneth~J Boote, Belay~T Kassie, Willingthon Pavan, Vakhtang Shelia, Kwang~Soo Kim, Ixchel~M Hernandez-Ochoa, et~al.
\newblock A simple crop model.
\newblock \emph{European Journal of Agronomy}, 104:\penalty0 97--106, 2019.

\bibitem[Zhao et~al.(2020)Zhao, Queralta, and Westerlund]{zhao2020sim}
Wenshuai Zhao, Jorge~Pe{\~n}a Queralta, and Tomi Westerlund.
\newblock Sim-to-real transfer in deep reinforcement learning for robotics: a survey.
\newblock In \emph{2020 IEEE symposium series on computational intelligence (SSCI)}, pages 737--744. IEEE, 2020.

\bibitem[Zhou et~al.(2023)Zhou, Geng, Shen, Tao, Long, Lou, and Shen]{zhou2023thread}
Yucheng Zhou, Xiubo Geng, Tao Shen, Chongyang Tao, Guodong Long, Jian-Guang Lou, and Jianbing Shen.
\newblock Thread of thought unraveling chaotic contexts.
\newblock \emph{arXiv preprint arXiv:2311.08734}, 2023.

\bibitem[Zhou et~al.(2024)Zhou, Li, Wang, and Shen]{zhou2024visual}
Yucheng Zhou, Xiang Li, Qianning Wang, and Jianbing Shen.
\newblock Visual in-context learning for large vision-language models.
\newblock \emph{arXiv preprint arXiv:2402.11574}, 2024.

\end{thebibliography}
}


\end{document}